\def\year{2021}\relax
\documentclass[letterpaper]{article} 
\usepackage{aaai21}  
\usepackage{times}  
\usepackage{helvet} 
\usepackage{courier}  
\usepackage[hyphens]{url}  
\usepackage{graphicx} 
\usepackage{natbib}  
\usepackage{xcolor}
\usepackage{hyperref}
\usepackage{amsmath}

\usepackage{caption} 
\title{OCR Graph Features for Manipulation Detection in Documents}
\author{Hailey James, Otkrist Gupta, Dan Raviv \\}
\affiliations{
    hailey.james@lendbuzz.com, otkrist.gupta@lendbuzz.com, dan.raviv@lendbuzz.com \\

	Lendbuzz \\
    125 High Street \\
 	Boston, Massachusetts 02110 \\

}

\usepackage[switch]{lineno}

\begin{document}
\maketitle

\begin{abstract}
Detecting manipulations in digital documents is becoming increasingly important for information verification purposes. Due to the proliferation of image editing software, altering key information in documents has become widely accessible. Nearly all approaches in this domain rely on a procedural approach, using carefully generated features and a hand-tuned scoring system, rather than a data-driven and generalizable approach.  We frame this issue as a graph comparison problem using the character bounding boxes, and propose a model that leverages graph features using OCR (Optical Character Recognition). Our model relies on a  data-driven approach to detect alterations by training a random forest classifier on the graph-based OCR features. We evaluate our algorithm's forgery detection performance on dataset constructed from real business documents with slight forgery imperfections. Our proposed model dramatically outperforms the most closely-related document manipulation detection model on this task.
\end{abstract}

\section{Introduction}
The use of digital documents for information verification is becoming increasingly common in domains such as finance, insurance, and administration. For example, digital documents may be used for the purpose of filing insurance claims, verifying address or income, or expense management. As these use cases increase, opportunities for deception through document manipulation naturally follow, as successful forgery of these types of documents can lead to personal reward. Recent technological advances have made manipulation of digital documents easier than ever, allowing forgers to cheaply manipulate key portions of a document. These portions may include information such as account balance, address, name, or salary. Using simple image editing software, forgers can alter a few characters or words in the document to achieve their purposes. This type of forgery is particularly difficult to mitigate as these types of documents don't contain extrinsic artifacts such as watermarks for authentication purposes. Any method for detecting manipulation in these types of documents must thus rely on the intrinsic properties of the document. 

\begin{figure*}
  \includegraphics[width=1\textwidth]{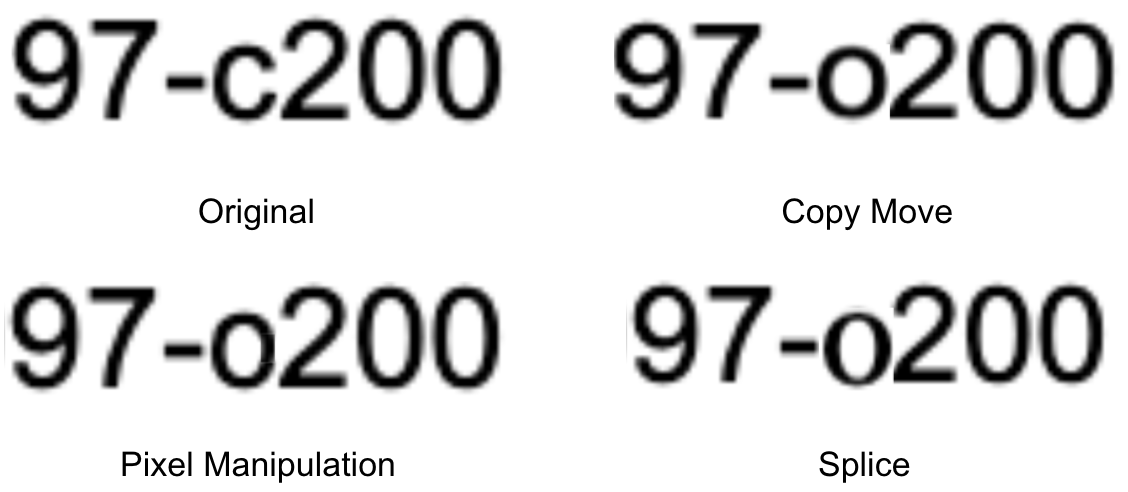}
\caption{Three possible manipulation types in document forgery include pixel manipulation, copy-move, and splicing. In pixel manipulation, individual pixels are manipulated to change the character. In this case, the gap in the 'c' is filled in to become an 'o'. In copy-move, the character is copied and pasted from another part of the document. In splicing, the character is copied and pasted from another document. In this case, font differences can be observed under careful inspection.}
\label{fig:three_manipulations}       
\end{figure*}

Digital document manipulation occupies a unique place within image manipulation detection, as properties such as color or texture that are normally leveraged for manipulation detection may not be as useful in digital documents. For example, many digital documents contain little or no color, and often an entirely white background. In addition, there are a variety of mechanisms available to alter a character or word in the document. The forger can copy and paste from another portion of the document, splice from a separate document, or construct the character from scratch by changing individual pixels (see Figure \ref{fig:three_manipulations}).

Manipulation detection methods for both images and documents rely on imperfections in the forgery process. In documents, these imperfections are often found in the alignment and sizing of the forged character \cite{van2011distortion}. We propose a method for detecting these forgery imperfections in document manipulation using OCR (Optical Character Recognition) graph features. We frame the problem of manipulation detection as a graph comparison problem, in which each character in the document is considered as the central node of a sub-graph within the larger graph of the document. Each sub-graph contains information about the central node and its neighbors, including features such as OCR box size, distance from the central node and alignment. We then train a random forest to recognize sub-graphs in which the central node has been manipulated. 

We construct a dataset which directly measures the ability of our model to detect these types of imperfections by stochastically shifting or scaling characters documents. Constructing this type of dataset constrains the model to rely on forgery imperfections, as texture or pixel-based features may be distorted or removed through post-processing like printing and scanning, Gaussian blurring or JPEG compression \cite{bayar2016}.

Our contributions can be summarized as follows:
\begin{itemize}
\item We are the first to consider the problem of document manipulation as a graph comparison problem and to leverage OCR features as graph features for manipulation detection. Our proposed method significantly outperforms the most closely related model designed for document manipulation detection.
\item We present a data-driven approach that uses a random forest architecture,  in contrast with the procedural approaches previously employed for detecting forgeries.
\item We evaluate our algorithm on a dataset that directly measures the ability of a manipulation detection model to detect forgery imperfections.
\end{itemize}

\section{Related Work}
\subsection{Manipulation Detection in Images}
Digital document manipulation detection occupies a space within image manipulation detection more broadly and builds upon work in this area. For example, \cite{fridrich_2003} investigate detection of copy-move forgery in images using block matching methods. \cite{lin_2009} develop a method for detecting copy-move forgery in images using Hu moment features on blocks of pixels. \cite{popescu2005exposing} use a pixel-based method for resampling using the statistical artifacts created during the manipulation operation.  \cite{stamm2010forensic} use a similar approach for detecting contrast enhancement, histogram equalization and JPEG-compression.

However, documents present unique challenges and opportunities when compared with manipulation detection in images. For example, as noted by \cite{lin_2009}, one challenge in detecting copy-move forgeries in images is trying to compare every possible pair of pixel blocks. In contrast, in documents this number of comparisons can be drastically reduced by ignoring white space and only comparing boxes containing the same alphanumeric character. Additionally, methods designed for forgery detection in images that leverage texture or pixel-based statistical properties may perform worse on documents due to large amounts of white space and less color information, as well as the potential for printing and scanning documents and removing these features.  Our proposed model in particular attempts to leverage the structured and predictable nature of documents to develop more discriminative features.

\subsection{Manipulation Detection in Documents}
Several methods for detecting forgeries rely on identifying the type of printer or exact machine from which a document was printed \cite{van2013automatic, shang2014detecting}. In contrast, \cite{ahmed_2014} use intrinsic document elements, or portions of the document that remain the same across different copies of the document, to detect manipulations in documents from the same source. \cite{bertrand_2015} employ a conditional random field model and font features to detect manipulations in which a character or word has been spliced from a document with different fonts. \cite{abramova_2016} investigate methods for detecting copy-move forgery in scanned text documents using block-based methods on characters that have been copied onto a "forgery" line at the bottom of the page. \cite{beusekom_2013} develop a method using text-line alignment and skew features to determine the authenticity of each line on the page, in contrast with other works (including ours) that examine alterations on a word or character level. \cite{shang_2015} use geometric properties of characters to distinguish between documents printed from laser printers, ink printers, and electrostatic copiers for purposes of document authentication. \cite{cruz_2017} use Otsu binarization to extract patches to compare using local binary pattern features combined with a Support Vector Machine (SVM) for forgery classification.  \cite{beusekom_2015} present findings from deploying manipulation techniques on real-world datasets to identify weaknesses. \cite{chernyshova_optical_2019} propose a system for optical font recognition for the purpose of forgery detection. \cite{sidere_2017} publish a dataset for evaluating performance on forgery detection in French payslips. \cite{gupta_2020} explore the benefits of using boosting and bagging in forgery detection. 

\begin{figure*}
\centering
  \includegraphics[width=1\textwidth, angle=270, scale=1.3]{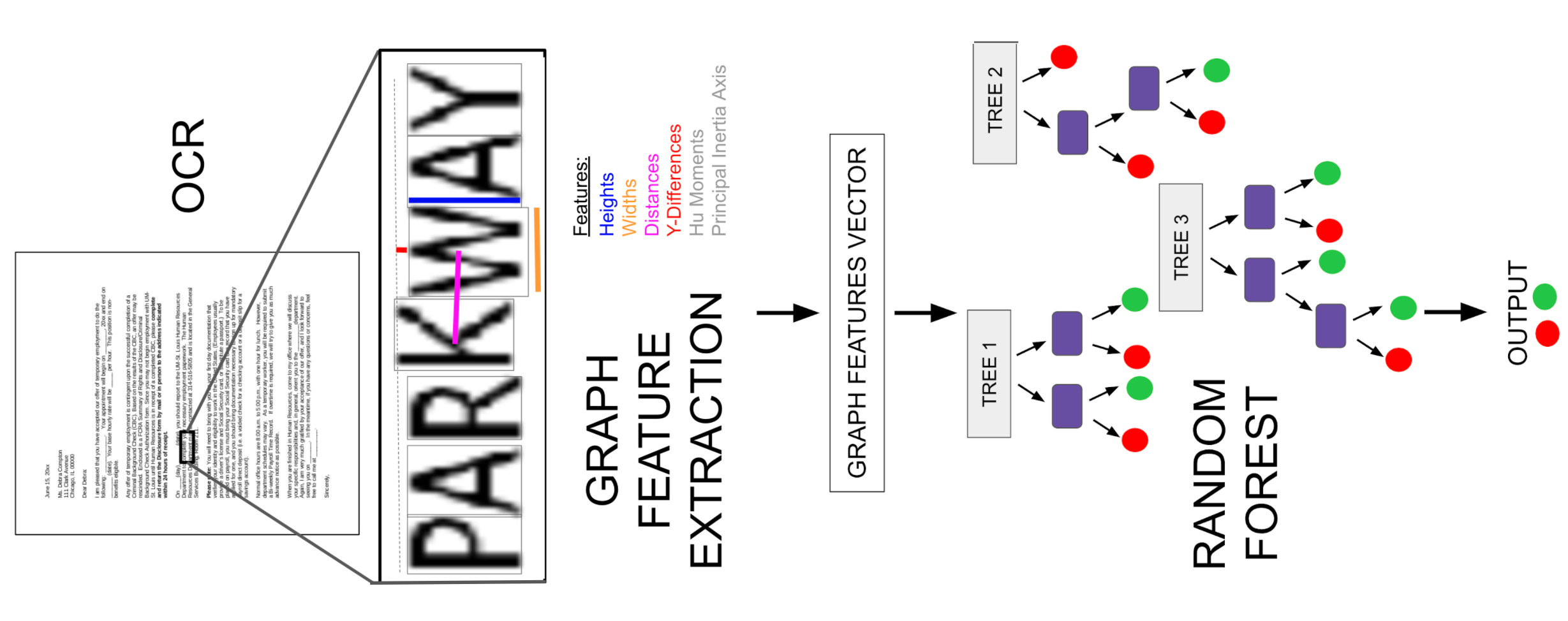}
\caption{Our pipeline begins with extracting character bounding boxes using Optical Character Recognition (OCR) via Tesseract. The boxes are then used to construct graph features for each potentially manipulated character. In this figure, the 'K' is the character being examined for manipulation. Hu moments and principal inertia axis features are not shown. The graph features are then inputted into a trained random forest, which classifies the character as manipulated or pristine.}
\label{fig:pipeline}   
\end{figure*}

\section{Methodology}

We frame the manipulation detection problem in documents as a graph comparison problem, where each character is the center of a graph containing $2n+1$ nodes, consisting of the central character node and $n$ nodes on each side horizontally. We experiment with various values of $n$, ranging from 3-9 (see Table \ref{tab:rf_hps}). We attempt to include only nodes that are on the same line of text by measuring the difference in the y-values of each node in comparison with the heights. Experimentally, we found that when the difference between the y-values of the top of the OCR boxes was less than $.85$ of the height of the taller box, the boxes were on the same line. In other words, nodes $n1$ and $n2$ are considered to be on the same line when 

\begin{equation}
|y0_{n1} - y0_{n2}| < (.85 * max(height_{n1}, height_{n2})
\end{equation}

where $y0_n$ indicates the y-value of node $n$ and $height_n$ corresponds to the height of node $n$. When the central node fell near the left or right end of a line, we imputed the missing values with the values from the other side of the central node when possible. 

Each of the features of each node is concatenated into a single vector to describe the sub-graph for each character in the document. For training, we assign the label $1$ when the central node has been manipulated, and $0$ when the central node is pristine or unaltered. We then train a random forest on the dataset.

The pseudo-code for reproducing our model is provided in the supplementary materials.

\begin{figure*}
  \includegraphics[width=1\textwidth]{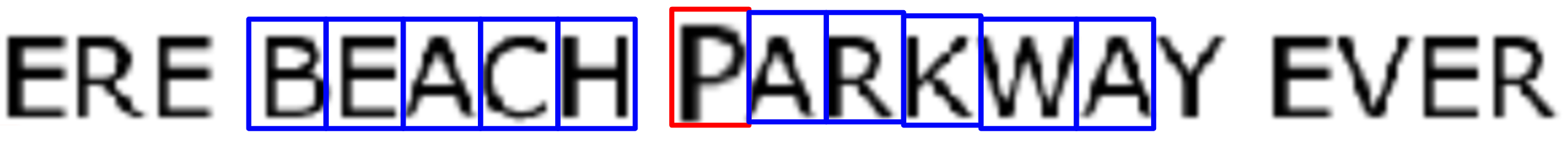}
\caption{The sub-graph contains the central node (red, 'P'), or the OCR character box being inspected for manipulation artifacts, and $n$ neighbors on either side (blue). Features include box characteristics such as height, width, respective y-value, distance from the central node, Hu moments, and principal inertia axis.}
\label{fig:features}       
\end{figure*}

\subsection{Features}
For each node in the sub-graph, we construct a set of features to describe the node, including:
\begin{itemize}
\item Height and width: Tesseract OCR provides the bounding box of each character, from which the approximate height and width of the character can be determined. If a character has been manipulated, imperfections in the manipulation often manifest in a slight character size difference.
\item Y-value Difference: We take the relative displacement of neighbouring nodes along the y-axis. We consider the uppermost y-value of the central node as 0, and compute the positive or negative difference of the neighboring node y-values with respect to the central node. This feature is useful as a proxy for alignment error, as exact alignment of the forged character is difficult to achieve even with digital image manipulation tools.
\item Distance: The Euclidean distance from the center of the each node to the center of the central node is calculated. Information about distance from the central node allows greater weighting of nodes that are closer to the central node. The distance is formulated as

\begin{equation}
d((x_1,y_1), (x_2, y_2)) = \sqrt{(x_1-x_2)^2 + (y_1-y_2)^2}
\end{equation}

where $(x_1,y_1), (x_2, y_2)$ refer to the center of the central node character box and the center of a neighboring node character box respectively.

\item Hu Moments: The seven Hu moments are invariate to rotation, scale and translation and allow for information about the character value within the box \cite{hu_moments, hu_analysis}. The seven Hu moments are defined as 

\begin{equation}
\begin{split}
    & \phi_{1} = \eta_{20} + \eta_{02}\\
    & \phi_{2} = (\eta_{20} - \eta_{02})^2 + 4\eta_{11}^2\\
    & \phi_{3} = (\eta_{30} - 3\eta_{12})^2 + (3\eta_{21} - \mu_{03})^2\\
    & \phi_{4} = (\eta_{30} - \eta_{12})^2 + (\eta_{21} - \mu_{03})^2\\
    & \phi_{5} = (\eta_{30} - 3\eta_{12})(\eta_{30} + \eta_{12})[(\eta{30} + \eta{12})^2 - 3(\eta_{21}+\eta_{03})^2]\\
    &+(3\eta{21}-\eta_{03})(\eta_{21}+\eta_{03})[3(\eta_{30} + \eta_{12})^2 - (\eta{21}+\eta{03})^2]\\
    & \phi_{6} = (\eta_{20}-\eta{02})[(\eta_{30}+\eta{12})^2-(\eta_{21}+\eta{03})^2]\\
    &+4\eta_{11}(\eta_{30}+\eta_{12})(\eta_{21}+\eta{03})\\
    & \phi_{7} = (3\eta_{21}-\eta_{03})(\eta_{30}+\eta_{12})[(\eta_{30}+\eta{12})^2-3(\eta_{21}+\eta{03})^2]\\
    & -(\eta_{30}-3\eta_{12})(\eta_{21}+\eta_{03})[3(\eta_{30}+\eta_{12})^2-(\eta_{21}+\eta_{03})^2]\\
\end{split}
\end{equation}

where each normalized centroid moment $\eta_{pq}$ is defined as 

\begin{equation}
    \eta_{pq} = \frac{\mu_{pq}}{\mu_{00}^\gamma}, \gamma=(p+q+2)/2,p+q=2,3,...
\end{equation}

and each centroid moment $\mu_{pq}$ is is computed as

\begin{multline}
\mu_{p,q} = =\int_{-\infty}^{\infty} \int_{-\infty}^{\infty}(x-\overline{x})^p(y-\overline{y})^q f(x,y)dx,dy
\\
p,q = 0,1,2...
\end{multline}
where the pixel point $(\overline{x}, \overline{y}$) is the centroid of the image, or in our case, the centroid of the character bounding box.
\item Principal Inertia Axis: The Principal Inertia Axis is obtained through Singular Value Decomposition (SVD) on the $\mu_{20}$ , $\mu _{11}$ and $\mu _{02}$ Hu  moments.  
\end{itemize}

\section{Experiments}

As described above, we construct a random forest using Python \cite{van1995python} and the sci-kit learn library \cite{scikit-learn}. We ran randomized hyper-parameter search for 480 iterations on search space of the parameters listed in Table \ref{tab:rf_hps}. Each experiment was run on a cross-validation set of five folds, and the average performance and standard deviation of the folds are reported in the Results section. The experiments were performed on an Intel Core i7-8700 machine with x86-64 architecture and 12 cores.

\begin{table}
\begin{tabular}{ll}
Number of trees                                                               & \begin{tabular}[c]{@{}l@{}}1000, 250, 1500, 1750, 2000, \\ 2250, 2500, 2750, 3000, 3250, \\ 3500, 3750, 4000, 5000\end{tabular} \\ \hline
Max tree depth                                                                & \begin{tabular}[c]{@{}l@{}}5, 10, 15, 20, 25, \\ 30, 35, 40, 45, 50\end{tabular}                                                 \\ \hline
\begin{tabular}[c]{@{}l@{}}Minimum number of \\ samples per leaf\end{tabular} & 4, 6, 8, 10, 12    \\ \hline   
\begin{tabular}[c]{@{}l@{}}Number of Neighbors ($n$)\end{tabular} & 3, 5, 7, 9 
\end{tabular}
\caption{Hyper-parameter ranges for the proposed model}
\label{tab:rf_hps}
\end{table}

\subsection{Dataset}

Our dataset consists of 359 finance-related documents, including bank statements, offer letters, credit card statements, bills, and tax returns. We split the documents into train and test sets, with 287 and 72 documents in each set respectively. Each document has one or more pages, together totalling 1470 pages for training and 389 pages for testing.

Each document original is a PDF, for which the PDF text boxes are stored in the document. The PDF is first converted to an image, and as we iterate through the character boxes in the document, we alter the box on the image through either shifting or scaling with probability $0.05$. Scaling or shifting is applied stochastically within four ranges of values: shifting 1-5px, shifting 5-10px, scaling 7\%-14\%, and scaling 15\%-25\%. Empirically, we found that shifting or scaling by values smaller than these did not change the size or position of the character. Values larger than these produces manipulations that were easily visible to the human eye and often significantly disrupted the OCR extraction. 

\begin{figure*}
  \includegraphics[width=1\textwidth]{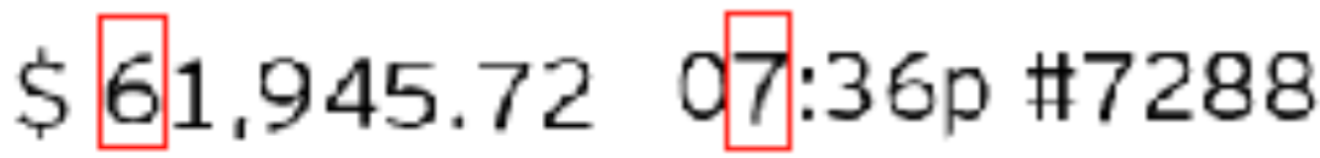}
\caption{An example of shifting (left) and scaling (right) manipulations. On the left, the six has been shifted up slightly to mimic the y-value alignment artifact introduced during a copy-move, pixel adjusting, or splicing operation. On the right, the seven has been slightly enlarged, as could be similarly expected during manipulation operations.}
\label{fig:shift_scale}       
\end{figure*}

If the box is altered, we store the coordinates of the altered box in the image for later use as ground truth. The resulting dataset is a collection of PNG images of documents of which $~5\%$ of the characters have been altered through scaling or shifting based on the PDF text boxes. Notably, the PDF text boxes are not provided to the manipulation detection model, and are distinct from the character boxes obtained through Tesseract OCR. See Figure \ref{fig:shift_scale} for examples of scaling and shifting.

While the original documents used for experimentation cannot be shared due to their sensitive nature, we provide pseudo-code in the supplementary materials for processing the dataset which can be performed on any set of similar real-world business documents. 

\subsection{Comparison} \label{comparison}

We compare our model's performance with the system proposed in \cite{bertrand_2013}, which similarly uses intrinsic document features on the character level to detect character-level manipulations. The proposed method also uses Tesseract OCR \cite{tesseract} for character extraction. The authors combine the results of two separate methods for classification of characters as manipulated or pristine.

In the first method, the authors  compute a distance between characters of the same alphanumeric type by constructing feature vectors consisting of the seven Hu moments \cite{hu_moments, terrades_2007} and calculating the Euclidean distance. Character pairs are flagged when the distance is either suspiciously small, indicating a copy-move forgery, or suspiciously large, indicating a character that is of a different size or font than the other characters of the same type in the document. 

In the second method, the authors propose finding a set of forgery clues grouped by a vector $W$ for each character. The feature vector $W$ consists of the character size based on the bounding box as obtained through Tesseract OCR, the character principal inertia axis as obtained through Singular Value Decomposition (SVD) on the $\mu_{20}$ , $\mu _{11}$ and $\mu _{02}$ Hu  moments, and the character horizontal alignment on the line level. The feature vector $W$ of each character is then compared to a data model $M_c$ for that character alphabetic class using the Mahalanobis distance, which is generated with the document training set. Similar to the first method, a scoring system determines if the distance is small enough or large enough to arouse suspicion of manipulation. 

Finally, the two indicators from the two methods are combined to give each character a score, for a final classification of manipulated or pristine. 

This method is an appropriate comparison for our method due to its similar character-based features and reliance on Tesseract OCR for determining character bounding boxes. Specifically, the feature extraction methods used by our model are a subset of those used by the  model proposed in \cite{bertrand_2013}. Comparison with this model allows us to disentangle the utility of the feature extraction employed from the combined utility of the graph-based features and the random forest architecture. 

We performed random search on a range of values and tune the hyper-parameters for a total of 480 iterations. The ranges of each value are shown in Table \ref{tab:ogier_hps}. 

\begin{table}
\begin{tabular}{ll}
Method 1 Upper Threshold       & 80, 95, 90   \\ \hline
Method 1 Lower Threshold       & 0, 5, 10 \\ \hline
Mahalanobis Threshold          & 5, 6, 7   \\ \hline
Log Transform Hu Moment Values & True, False     
\end{tabular}
\caption{Hyper-parameter ranges used for training the model described in \cite{bertrand_2013}}
\label{tab:ogier_hps}
\end{table}

\section{Results}
The results in Tables \ref{tab:scale_big}, \ref{tab:scale_small} \ref{tab:shift_big}, \ref{tab:shift_small} indicate that while the model proposed in \cite{bertrand_2013} achieves a reasonable F1 score and high recall, it reports a high number of false positives. In contrast, our model is able to achieve impressive performance given even minute manipulations such as shifting by a single pixel or scaling by as little as 7\%. We expected a higher F1 score with the "easier" or larger manipulations than was achieved (see Tables \ref{tab:scale_big} and \ref{tab:shift_big}), and hypothesize that larger manipulations may disrupt our models ability to correctly distinguish between characters on the same line and characters on different lines.
\begin{table}
\begin{tabular}{lllll}
          & \begin{tabular}[c]{@{}l@{}}Proposed \\ Method\end{tabular} & STD          & \begin{tabular}[c]{@{}l@{}}Bertrand \\ 2013\end{tabular} & STD          \\ \hline
Precision & \textbf{0.8651}                                            & $\pm$ 0.0134 & 0.2927                                                   & $\pm$ 0.0144 \\
Recall    & 0.7461                                                     & $\pm$ 0.0181 & \textbf{0.7862}                                          & $\pm$ 0.0170 \\
Accuracy  & \textbf{0.9857}                                            & $\pm$ 0.0112 & 0.3790                                                   & $\pm$ 0.0128 \\
F1 Score  & \textbf{0.8012}                                            & $\pm$ 0.0181 & 0.4265                                                   & $\pm$ 0.0170
\end{tabular}
\caption{\label{tab:scale_big}Mean metrics and standard deviations for scaling 5\% of characters by a range of 15\% to 25\% for each of the five splits}
\end{table}

\begin{table}
\begin{tabular}{lllll}
          & \begin{tabular}[c]{@{}l@{}}Proposed \\ Method\end{tabular} & STD          & \begin{tabular}[c]{@{}l@{}}Bertrand \\ 2013\end{tabular} & STD          \\ \hline
Precision & \textbf{0.8709}                                            & $\pm$ 0.0143 & 0.2495                                                   & $\pm$ 0.0153 \\
Recall    & 0.6177                                                     & $\pm$ 0.0221 & \textbf{0.7971}                                          & $\pm$ 0.0315 \\
Accuracy  & \textbf{0.9817}                                            & $\pm$ 0.0212 & 0.3506                                                   & $\pm$ 0.0122 \\
F1 Score  & \textbf{0.7228}                                            & $\pm$ 0.0210 & 0.3800                                                   & $\pm$ 0.0206
\end{tabular}
\caption{\label{tab:scale_small}Mean metrics and standard deviations for scaling 5\% of characters by a range of 7\% to 14\% for each of the five splits}
\end{table}

\begin{table}
\begin{tabular}{lllll}
          & \begin{tabular}[c]{@{}l@{}}Proposed \\ Method\end{tabular} & STD          & \begin{tabular}[c]{@{}l@{}}Bertrand \\ 2013\end{tabular} & STD          \\ \hline
Precision & \textbf{0.8187}                                            & $\pm$ 0.0155 & 0.2111                                                   & $\pm$ 0.0166 \\
Recall    & 0.7440                                                     & $\pm$ 0.0162 & \textbf{0.8047}                                          & $\pm$ 0.0161 \\
Accuracy  & \textbf{0.9854}                                            & $\pm$ 0.0172 & 0.3318                                                   & $\pm$ 0.0146 \\
F1 Score  & \textbf{0.7796}                                            & $\pm$ 0.0102 & 0.3343                                                   & $\pm$ 0.0106
\end{tabular}
\caption{\label{tab:shift_big}Mean metrics and standard deviations for shifting 5\% of characters by a range of 5-10 pixels for each of the five splits}
\end{table}

\begin{table}
\begin{tabular}{lllll}
          & \begin{tabular}[c]{@{}l@{}}Proposed \\ Method\end{tabular} & STD          & \begin{tabular}[c]{@{}l@{}}Bertrand \\ 2013\end{tabular} & STD          \\ \hline
Precision & \textbf{0.8604}                                            & $\pm$ 0.0101 & 0.2498                                                   & $\pm$ 0.0091 \\
Recall    & 0.6666                                                     & $\pm$ 0.012  & \textbf{0.7896}                                          & $\pm$ 0.014  \\
Accuracy  & \textbf{0.9829}                                            & $\pm$ 0.0198 & 0.3494                                                   & $\pm$ 0.0111 \\
F1 Score  & \textbf{0.7512}                                            & $\pm$ 0.0121 & 0.3795                                                   & $\pm$ 0.0116
\end{tabular}
\caption{\label{tab:shift_small}Mean metrics and standard deviations for shifting 5\% of characters by a range of 1-5 pixels for each of the five splits}
\end{table}
			
\section{Conclusion}
We present a system for detecting forgery imperfections using OCR graph features with a random forest architecture. We evaluate this algorithm on detecting small shifting and scaling manipulations in real financial documents. As shown, our model outperforms the most closely related model significantly on this task. It seems that at least for documents, a data-driven approach may be required instead of a procedural approach for obtaining satisfactory manipulation detection results. In addition, it appears that leveraging the structured nature of digital documents by incorporating graph features offers additional performance gains. The unique nature of digital documents, and their importance in manipulation detection research, warrants special attention and inquiry into developing methods for identifying malicious manipulations.

Future work could include leveraging graph anomaly detection methods for a more efficient graph-based detection system. Additionally, the OCR graph features could be combined with pixel-based features from traditional image manipulation detection for potential improved performance. We also observed that the character boxes resulting from Tesseract OCR were sometimes inaccurate, failing to accurately enclose each character. Improving the underlying OCR method could further improve the manipulation detection results.

\section{Ethics Statement}

Fraud detection is an active area of research and can be of great benefit to society. Detection and prevention of document fraud can potentially help reduce misinformation and improve resource allocation. Users and organizations often spend a large amount of resources trying to mitigate fraud, usually through human auditors. Automating fraud detection can help reduce this burden. However, as with any machine learning model, caution should be exercised, particularly when considering acceptable false positive rates. Additional human verification may be required when attempting to detect fraud in real life scenarios. Data driven approaches like ours can help mitigate such risks but may not eliminate them, and further research in this domain should be pursued.

\bibliography{ocr_gf.bib}
\end{document}


\maketitle

\section{Proposed Model Algorithm}

\section{Dataset Construction Algorithm}

\begin{algorithm}[]
 \KwData{business PDF documents, split into train and test}
 \KwResult{PNG images with slight shifting and scaling character-level forgery imperfections}
 initialization\;
 \For 
 \While{not at end of this document}{
  read current\;
  \eIf{understand}{
   go to next section\;
   current section becomes this one\;
   }{
   go back to the beginning of current section\;
  }
 }
 \caption{How to write algorithms}
\end{algorithm}